\documentclass{article}
\usepackage{fullpage}
\usepackage{natbib}
\usepackage{amsmath}
\usepackage{amsfonts}
\usepackage{hyperref}
\usepackage{graphicx}
\usepackage{authblk}
\usepackage[framemethod=TikZ]{mdframed}

\usepackage{titling}
\newcommand{\ptr}{P^\text{tr}}
\newcommand{\pte}{P^\text{te}}
\newcommand{\hash}[1]{\texttt{hash(#1)}}
\newcommand{\public}{\texttt{model\_public}}
\newcommand{\private}{\texttt{model\_private}}

\hypersetup{
    colorlinks,
    linkcolor={blue!50!black},
    citecolor={blue!50!black},
    urlcolor={blue!50!black}
}

\mdfsetup{%
   middlelinecolor=blue!30,
   middlelinewidth=1pt,
   backgroundcolor=blue!10,
   skipabove=0.2cm,
   innertopmargin=4pt,
   innerbottommargin=5.5pt,
   innerleftmargin=8pt,
   innerrightmargin=8pt,
   roundcorner=2pt}

\title{Measuring and signing fairness as performance\\ under multiple stakeholder distributions}
\author{David Lopez-Paz, Diane Bouchacourt, Levent Sagun, Nicolas Usunier}
\affil{Meta AI, Paris, France}
\date{\vspace{-5ex}}

\begin{document}

\maketitle

\begin{figure}[h!]
  \begin{center}
  \includegraphics[width=\textwidth]{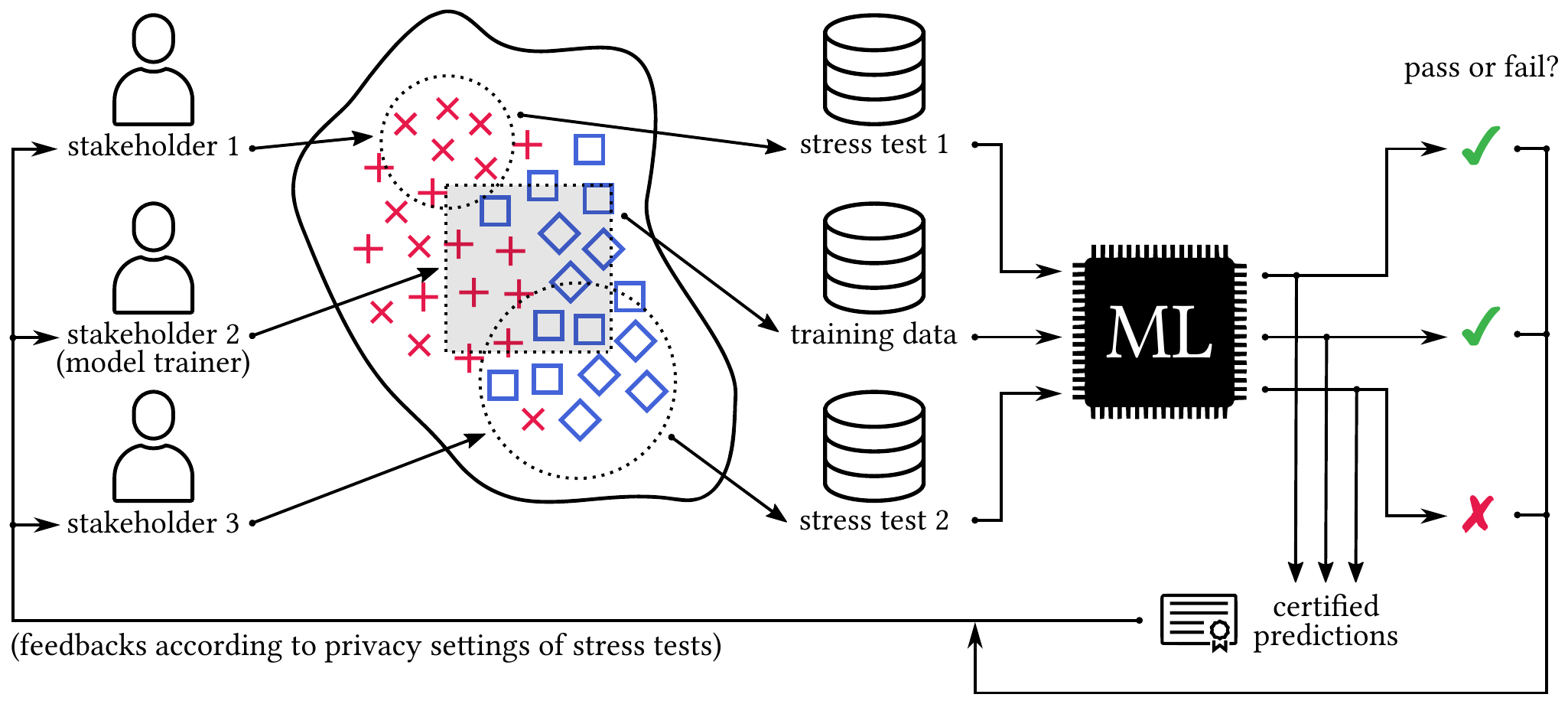}
  \end{center}
  \caption{The proposed framework of \emph{stress tests}. Stakeholders curate \emph{stress tests} containing examples that illustrate their (possibly competing) interests. The machine learning system passes or fails each stress test, informing the relevant stakeholders on how to improve the fairness of the system. This feedback flows according to the privacy settings of each stress test to avoid, for instance, their use as additional training data. The system provider digitally signs their predictions, assuming a degree of accountability.}
  \label{fig:mainfig}
\end{figure}

\begin{abstract}
  As learning machines increase their influence on decisions concerning human lives, analyzing their fairness properties becomes a subject of central importance.
  Yet, our best tools for measuring the fairness of learning systems are rigid fairness metrics encapsulated as mathematical one-liners, offer limited power to the stakeholders involved in the prediction task, and are easy to manipulate when we exhort excessive pressure to optimize them.
  To advance these issues, we propose to shift focus from shaping fairness metrics to curating the distributions of examples under which these are computed.
  In particular, we posit that every claim about fairness should be immediately followed by the tagline ``\emph{Fair under what examples, and collected by whom?}''.
  By highlighting connections to the literature in domain generalization, we propose to measure fairness as the ability of the system to generalize under multiple \emph{stress tests}---distributions of examples with social relevance.
  We encourage each stakeholder to curate one or multiple stress tests containing examples reflecting their (possibly conflicting) interests.
  The machine passes or fails each stress test by falling short of or exceeding a pre-defined metric value.
  The test results involve all stakeholders in a discussion about how to improve the learning system, and provide flexible assessments of fairness dependent on context and based on interpretable data.
  We provide full implementation guidelines for stress testing, illustrate both the benefits and shortcomings of this framework, and introduce a cryptographic scheme to enable a degree of prediction accountability from system providers.
\end{abstract}

\section{Introduction}
\label{sec:intro}

Machine learning is permeating critical social sectors such as banking~\citep{hand1997statistical}, justice~\citep{dressel2018accuracy}, citizen services~\citep{mehrabi2021survey}, government~\citep{engstrom2020government}, police~\citep{perry2013predictive, lum2016predict}, and the public sector~\citep{berryhill2019hello}.
Given the soaring impact of these systems on billions of people, it is imperative to monitor and enforce \emph{fairness} in their automated decisions concerning human lives~\citep{fairmlbook}.
Namely, when do machine learning systems provide an impartial treatment to individuals and do not discriminate between social groups such as those defined by socio-economic status, race, or gender \citep{mittelstadt2016ethics, tsamados2021ethics}?

Defining fairness is an admittedly difficult task that depends heavily on the context at hand.
Currently, our main tools to measure fairness in machine learning are \emph{fairness metrics}---mathematical one-liners probing specific relationships between predictions and the variables in data~\citep{verma2018fairness}.
As one example, the metric of \emph{independence} promotes classification systems with equal outcomes across across groups categorized by sensitive attributes.
While useful, an excessive focus on enforcing metrics places three major roadblocks towards the study of fairness.
(I) These metrics are rigid, since they are simply statements about statistical dependencies in data~\citep{fairmlbook}.
While we are witnessing an incredible research effort to grow the index of available metrics, it is unlikely that a rule-based system will ever capture the nuance and sophistication necessary to delineate fairness.
(II) An exclusive focus on fairness metrics limits the participation of stakeholders.
This is because the two assets necessary to compute them, models and data, are often owned exclusively by system providers.
(III) Fairness metrics are easy to manipulate, so optimizing them blindly may lead to unfair classification systems~\citep{corbett2018measure}.
For instance, consider enforcing the metric of independence when designing a system to predict the probability of a defendant re-offending.
Tossing a coin at each court case would satisfy the independence metric unjustly, while actively enforcing it would punish women defendants excessively~\citep{corbett2018measure}.
Generally speaking, we shall always bear in mind the uncanny ability of learning machines to find unexpected shortcuts to satisfy any fixed metric imposed on them~\citep{geirhos2020shortcut}.

Central to the sequel is to argue that a focus on metrics distracts our attention from the distributions of data under which these metrics are computed.
Because of all mentioned above, this work advocates to shift our focus from a parametric rule-based to a non-parametric data-based practice to evaluate the fairness of machine learning systems.
Our approach borrows inspiration from the research literature in \emph{domain generalization}~\citep{gulrajani2020search}, where fairness metrics were independently discovered and purposed as regularizers to improve the out-of-distribution performance of classifiers.
Which domain generalization regularizer best suits a task depends on the causal structure of the data, in most cases unknown.
Therefore, researchers in domain generalization opted for a data-driven approach to measure out-of-distribution performance: simply measure classification accuracy over novel distributions of examples.
Borrowing these ideas into the realm of fairness,
\begin{mdframed}
  We propose to measure the fairness of a machine learning system as its ability to generalize across multiple \emph{stress tests} (societally-relevant distributions of examples) collected by diverse stakeholders for the task at hand.
\end{mdframed}
It is our view that every claim about the fairness of a machine learning system should be immediately followed by the tagline:
``\emph{Fair under what examples, and collected by whom?}''.

Figure~\ref{fig:mainfig} summarizes the proposed framework of \emph{stress tests}.
Foremost, we encourage the stakeholders involved in the learning task to curate one or more stress tests of examples to illustrate their possibly competing interests.
Stakeholders include model builders (providing the training data), responsible AI and ethics teams, activists, expert witnesses in the court of law, journalists, groups, and individuals.
Each stress test is akin to an ``unit test'' in software development, evaluating the performance of the learning system under controlled experimental conditions----here, under different distributions of examples.
A machine learning system passes a stress test if it achieves or exceeds a predefined metric value (e.g., classification performance).
Opting for a conditional definition of fairness, we say that the system is \emph{fair with respect to a collection of stress tests} if it passes all the stress tests in the collection.
To enable accountability, we advocate that system providers digitally sign their predictions.
Stress tests complement rule-based systems of fairness measurement along three pillars.
 
\paragraph{(I) Stress tests are a flexible tool to measure fairness.}
The framework accommodates context easily, since stakeholders collect multiple societally-relevant distributions of examples anew for each learning task.
Stress tests go beyond ``sweeping over the values of the sensitive attribute'', since they can differ due to general shifts in data distribution.
They can illustrate changes in difficult-to-describe environmental conditions, unstable correlations, hidden confounders, causal interventions~\citep[c.f. counterfactual fairness]{kusner2017counterfactual}, or difficult to define (e.g., not necessarily discrete, overlapping) social groups.
All of this is possible without the need to include sensitive attributes in the stress tests.
As opposed to parametric rule-driven approaches, stress tests are a non-parametric data-driven solution to assess fairness.

\paragraph{(II) The framework of stress testing involves all stakeholders properly.} Stress testing distributes the power associated to data, the most important asset in machine learning.
Stress testing establishes clear-cut duties: everyone is responsible for the curation of reliable stress tests, and system providers are responsible for providing and signing predictions.
The framework embraces subjectivity, since every stakeholder may express different---even incompatible---views through their stress tests.
At a higher level, measuring performance on multiple distributions is a rich language to align the incentives of humans and machines~\citep{radford2021learning}.
Each stress test is a different perspective on human behaviour, a key source to learn about human preference~\citep{russell2019human}.
On the one hand, the more stress tests a classifier passes, the more confidence human operators may place on the system.
While human-like performance does not imply human-like strategy, it is increasingly difficult for machines to be right for the wrong reason when passing a larger amount of diverse stress tests.
In contrast, passing a fairness metric on a given distribution says nothing about that metric in other distributions.
On the other hand, failing a stress test is a call to reconsider our data, improve the machine learning system, reduce the scope of the problem, or question if the learning task belongs to the realm of machine learning at all.
Failing a stress test is an honest reminder about the ever-present gap between the human understanding of a task and the machine's shortcut to a solution.

\paragraph{(III) Stress testing is more difficult to manipulate than fairness metrics.}
This is particularly the case when forbidding to add stress examples to the training data, or when some stakeholders (e.g., ethics teams) hold their stress tests private to the trainers of the system.
These privacy mechanisms implement an interface similar to a private Kaggle leaderboard~\citep{blum2015ladder}, where system providers cannot overfit or find shortcuts to pass the fairness stress tests.
On the flip side, it is helpful to show individual mistakes to system providers so they can better understand the types of mistakes that the system makes, and how to fix them.

Revealing unfairness in machine learning systems by subjecting them to difficult testing examples is increasingly common~\citep{klare2015pushing, levi2015age, ImageNetRoulette, shankar2017no, de2019does, karkkainen2019fairface, karkkainen2019fairface, hazirbas2021towards, radford2021learning, schumann2021step, dollar}.
The literature in fairness knows well the complexity and importance of building diverse datasets fairly representing social groups~\citep{drosou2017diversity, chasalow2021representativeness}.
For instance, the large-scale survey of~\citet{holstein2019improving} about enforcing fairness highlighted ``the central importance of careful test set design to detecting potential fairness issues'' and considers ``extremely useful'' to ``support practitioners in collecting and curating high-quality datasets in the first place, with an eye towards fairness in downstream ML models''.
The practice of stress testing is also related to internal audits~\citep{raji2020closing} and the curation of archives by diverse stakeholders~\citep{jo2020lessons}.
Our contribution is to provide perspective, sharpen our language, and establish useful links to the domain generalization literature.
We embed out-domain evaluation into a complete framework, making the choice of data explicit in the specification of the machine learning system. This enables the development of data-driven mitigation strategies to end ``metric fights'' over distributions of examples with dubious fairness properties, and takes a first step towards developing schemes that enable a degree of accountability from system providers.
That way, stress testing fosters discussions between stakeholders and system providers to improve the quality of service and redefine the operating regime of the system.
There is still much to do when it comes to distributing the power and authority of data across stakeholders, yet we believe a rigorous practice of stress testing moves the ball a few inches forward.
In words of \citet{mary}, during his invited talk at NeurIPS 2021:
\begin{quote}
  Because data has become so powerful it is imperative that we make it our professional collective responsibility to transfer the tools of data collection, aggregation, and sharing from engineers and the institutions that underwrite their work, wherever they might be, to the communities and members of society caring about the benefits and risks of what can be built.
\end{quote}

Next, Section~\ref{sec:tour} kicks-off the exposition with a quick tour on training and evaluating machine learning systems, as well as measuring their fairness according to simple metrics.
Section~\ref{sec:catalog} provides a catalog of real-world examples of claims about unfairness in classification systems.
Our intent with this catalog is to reveal a connection between what is popularly understood as unfairness and the issue of generalizing to distributions of examples different to those of the training data.
Section~\ref{sec:domain} leverages the previous catalog as a motivation to revisit the literature in domain generalization, which reveals a link between popular fairness metrics and their independent discovery as regularizers to improve out-of-distribution performance.
Section~\ref{sec:stress} introduces the framework of \emph{stress testing}, or how to measure fairness as the ability of a machine learning system to generalize across multiple and societally-important distributions of examples as provided by different stakeholders.
Section~\ref{sec:cert} proposes a cryptographic scheme for system providers to digitally sign their predictions, enabling a degree of accountability and allowing stakeholders to interact with each other in a trusted environment.
Finally, Section~\ref{sec:conclusion} offers some concluding remarks.

\section{A quick tour on learning fair classifiers}
\label{sec:tour}

We start by laying out some basic terminology that we will use during the remainder of our exposition.
Without loss of generality, we discuss the training and evaluation of binary classifiers, as well as the most common tools to measure their fairness.

\subsection{Training a binary classifier}

Consider the problem of learning a binary classifier.
To this end, we collect a training set of examples, each comprising a triplet $(x, a, y)$.
Here, $x$ is a vector of input features, $a$ is a sensitive attribute, and $y \in \{-1, +1\}$ is a binary label.
As it is often the case in machine learning practice~\citep{vapnik1992principles}, we assume that all training examples originate from some unknown training distribution $\ptr$.
Once our training data is available, we tune the parameters of the classifier so it produces appropriate prediction scores $r_i$ for each individual $i$ in the training set.
In particular, we ask our classifier to predict high scores when given input features $x_i$ associated to positive labels $y_i = +1$, and low scores otherwise.
These scores can be later calibrated (adjusted) to represent probabilities.

As a running example to settle our notation, let us consider the COMPAS decision support tool, designed by Northpointe Inc. and employed by U.S.~Courts to determine the likelihood of a defendant re-offending~\citep{compas}.
Here, training examples $(x, a, y)$ are a limited description of past defendants.
This description includes some input features $x$ (such as past criminal records, drug involvement), a sensitive attribute $a$ (such as race), and a binary target $y$ (in this case, $+1$ to indicate that the past defendant was re-offended, and $-1$ otherwise).
As stated above, training data stems from some bigger, underlying distribution.
In the sequel, let us assume that the training set is an incomplete list of past defendants from Florida's Broward County, which would serve as our training distribution $\ptr$.
Using this training data, we tune our classifier to predict high scores $r_i$ when given features $x_i$ of past defendants that did re-offend (labeled with $y_i = +1$), and low scores otherwise.

\subsection{Evaluating performance}

Once trained, it is important to assess the performance of our classifier on new, testing examples $(x', a', y')$.
Most machine learning applications assume that testing examples originate from the same distribution of the training examples~\citep{murphy2012machine}.
We call this evaluation \emph{in-domain}, and write the underlying assumption as $\pte = \ptr$, where $\pte$ is the distribution of testing examples.
Normally, the in-domain performance of a classifier increases as the amount of training data grows in size~\citep{vapnik1999nature}.
In contrast, we could collect testing examples from a different distribution $\pte \neq \ptr$.
In this case, we would be evaluating our classifier \emph{out-domain}.

Back to our running example, we can evaluate the performance of the COMPAS recidivism classifier by collecting a testing set of new cases of defendants.
On the one hand, when these testing cases concern Florida's Broward County, we are evaluating our classifier in-domain ($\pte = \ptr$).
On the other hand, when collecting these testing cases from a different state in the US, we are evaluating our classifier out-domain ($\pte \neq \ptr$).
Central to this manuscript is to argue that out-domain evaluation is tightly linked to measuring fairness---to this end, the sequel assumes issues regarding in-domain generalization addressed and negligible.

\subsection{Evaluating fairness}
\label{sec:rules}

Measuring the fairness of a classification system is commonly done in reference to some sensitive attribute $a$, which in this section (but not in stress testing) assumes part of all training and testing examples.
Different values of the sensitive attribute $a$ stratify data into societally-important groups (for instance, $a$ could index different genders).

The research community has proposed a myriad of measures and metrics to evaluate the fairness of machine learning systems~\citep{verma2018fairness}.
Three popular examples of fairness metrics are independence, separation, and sufficiency~\citep{fairmlbook}.
These are conditional independence statements between the random variables associated to the prediction scores $r$, the sensitive attribute $a$, and the label $y$.
First, \emph{independence} measures fairness as the level of statistical independence between the prediction scores $r$ and the sensitive attribute $a$.
Thus, a classification rule satisfies the independence metric when the distributions of scores are the same across values for the sensitive attribute.
Second, \emph{separation} measures fairness as the level of similarity of false negative and false positive errors across values of the sensitive attribute.
Therefore, \emph{separation} measures error in-domain.
Mathematically, this amounts to requiring that the prediction score $r$ and the sensitive attribute $a$ are conditionally independent given the label $y$.
Third, \emph{sufficiency} is tightly related to requiring calibrated prediction scores (representing probabilities) for every group defined by the sensitive attribute.
Formally, this is requiring that the label $y$ and the sensitive attribute $a$ are conditionally independent given the predictions score $r$.

Following with our example, an influential article in ProPublica~\citep{compas} raised issues regarding the fairness of the COMPAS recidivism classifier.
In their words, ``blacks are almost twice as likely as whites to be labeled a higher risk but not actually re-offend.
It makes the opposite mistake among whites: They are much more likely than blacks to be labeled lower risk but go on to commit other crimes''.
This is a claim of unfairness under the separation metric.
The developer of the COMPAS classifier, Northpointe Inc, issued a technical report to rebut ProPublica's analysis by arguing that the system satisfied other fairness metrics such as sufficiency~\citep{dieterich2016compas}.
While there was significant attention devoted to ``metric fighting'' throughout the scandal \citep{compasresponse,flores2016false,corbett2017algorithmic}, the discussion was disappointingly thin when it came to arguments about the distributions of examples under which these metrics are computed.
Yet, a perfect COMPAS classifier for Florida's Broward County passing any given fairness metric would likely fail it when evaluated on defendants from a different state.
For instance, the relationship learned by the classifier between postcode and recidivism from Broward County could revert for cases from Idaho's Canyon County (invented example), leading to wrong predictions---an issue of failing to generalize \emph{out-domain}.

\section{Unfairness in-the-wild: a catalog}
\label{sec:catalog}

We have just hinted at a possible relationship between out-domain generalization and fairness.
Before formalizing this link further, this section provides a list of examples of claims about unfairness found in the research literature and popular press.
The purpose of this ``catalog'' is two-fold.
On the one hand, we would like to build a better intuition about the subjectivity involved in defining fairness in different contexts.
On the other hand, we want to sharpen our sense of connection between the unfairness of a learning system and its failure to generalize out-domain.
To this end, for each entry in the catalog we include (i) a claim about unfairness concerning some underlying classifier, (ii) one or more references to the relevant literature and popular press, and (iii) one possible distribution of examples ``out-domain'' under which the classifier does not seem to generalize.
\begin{itemize}
  \item Algorithms to predict recidivism flagging non re-offending Black people twice as likely as white people~\citep{compas}.
  Non re-offending Black people.
  \item Commercial gender classification systems misclassifying Black women forty times more than white men~\citep{buolamwini2018gender}.
  Black women.
  \item Nikon cameras detecting Asian people as blinking~\citep{jozjozjoz}.
  Open-eyed Asian people.
  \item Object classifiers underperforming on images of lower-income households~\citep{de2019does} and Asian bridge-rooms~\citep{shankar2017no}.
  Low-income households and Asian bridgerooms.
  \item Google tagging photos of Black people with non-human labels~\citep{wired}.
  Black people.
  \item Facebook's video-caption matching system propagated a racially-sensitive tag from a past user-supplied text caption into a video showing a Black person~\citep{daily}.
  Black people.
  \item Automatic grading systems assigning lower scores to highly-qualified students belonging to minorities~\citep{ramineni2018understanding} or attending schools with a worse track record~\citep{ofqual}.
  High-performing students from minorities and lower-performing schools.
  \item Amazon's recruitment system disregarding highly-qualified woman candidates~\citep{amazoncv, amazoncv2}.
  Curriculum vitaes from high-performing woman candidates.
  \item Apple assigning low credit scores to highly-qualified woman applicants~\citep{applecredit}.
  Woman applicants with a strong credit record.
  \item YouTube's auto-captioning system underperforming for woman voices~\citep{youtube}.
  Audio recordings of woman speakers.
  \item Healthcare systems labeling sick Black patients with similar scores to healthy white patients~\citep{obermeyer2019dissecting}.
  Black patients with urgent healthcare needs.
  \item Toxicity prediction systems exhibiting a negative bias against minoritized speakers~\citep{sap2019risk}.
  Non-toxic tweets of African-American speakers.
  \item Sentiment prediction systems labeling sentences involving certain gender-race combinations with higher intensity~\citep{kiritchenko2018examining}.
  Neutral tweets from Black woman writers.
  \item Speech-to-text systems underperforming on certain accents~\citep{tatman2017gender} and African-American speakers~\citep{Koenecke7684}.
  Audio recordings of speakers from minorities and with foreign accents.
  \item Language identification systems underperforming on African-American English~\citep{blodgett2017racial}.
  Texts by African-American writers.
  \item Google Translate showing gender-role stereotypes when translating from gender-neutral languages such as Turkish~\citep{translate}.
  Turkish-English Sentences breaking gender-role stereotypes.
  \item Word embeddings showing gender-role stereotypes to a disturbing extent~\citep{bolukbasi2016man}.
  Texts to translate breaking gender-role stereotypes.
  \item LinkedIn classifying woman names as misspelled man names~\citep{linkedin}.
  Woman names.
\end{itemize}

Some claims about unfairness denounce the disparity of outcome between different groups of people~\citep{fairmlbook}.
We can also rephrase these claims as generalization issues.
In these cases, we must examine the performance of the classifier for \emph{two or more} datasets:
\begin{itemize}
  \item Google Search showing ads for arrest records when querying Black-sounding names~\citep{sweeney2013discrimination}.
  Datasets of white-sounding and Black-sounding names.
  \item Black people targeted twice as likely as whites by predictive policing algorithms~\citep{lum2016predict}.
  Datasets of Black and white people.
  \item White residents twice as likely than Black residents to qualify for Amazon's free same-day delivery~\citep{amazonsameday}.
  Lists of predominantly Black and white neighborhoods.
  \item Google Image Search shows men when querying for pictures of CEOs~\citep{imagesearch}.
  Datasets of query-image pairs with annotated gender.
\end{itemize}

Unfortunately the list goes on! It is evident that many of these examples reflect deep structural inequalities that are engraved in society. 
These examples can be dubbed as `tails' of distributions or `rare' examples which often reflect the choices of dataset builders~\citep{rostamzadeh2021thinking}.
Our aims with these examples are two-fold: (1) to emphasize the connection between proper stakeholder involement and engineering responsibility, and (2) to reveal how claims about unfairness in classification systems often have a rapport with issues of out-domain generalization.
Next, we deepen on the formal connection between fairness and out-domain performance to inspire an alternative way to rule-based fairness metrics.

\section{From fairness to domain generalization and back}
\label{sec:domain}

The previous catalog illustrates the multiple faces of fairness, and we would like to highlight three themes.
(I) The definition of fairness depends on context and is difficult to encapsulate into a mathematical one-liner.
(II) Claims about unfairness often reflect an attack against the interests of some stakeholders.
(III) Satisfying some definitions of fairness does not necessarily avoid undesirable classifiers~\citep{corbett2018measure}.
For instance, system creators can silence claims about disparate error rates by classifying at random or artificially sacrifying performance for some values of the sensitive attribute~\citep{corbett2018measure}.

Our best tools to date for measuring fairness in machine learning systems are simplistic, rigid metrics (see Section~\ref{sec:rules}).
Yet, in words of the British economist Charles Goodhart, ``when a metric becomes a target, it ceases to be a good metric''~\citep{strathern1997improving}.
This applies on a daily basis when dealing with learning machines, given their super-human ability to find unexpected shortcuts to satisfy any fixed metric imposed on them~\citep{geirhos2020shortcut}.
As long as these metrics do not capture the whole spectrum of human intention when it comes to defining fairness, we cannot control the side-effects of satisfying them mathematically.
Using a dramatic example, one effective strategy for an artificial intelligence to ``eradicate cancer'' is to kill all living beings~\citep{russell2019human}.

Historically, we find at least two warning calls on the perils of defining a complex concept using a finite set of simple rules.
On the one hand, building an ever-growing list of fairness metrics brought to halt a similar research effort in the 1960s and 1970s, and could already be having a detrimental effect on our research community~\citep{fifty}:
\begin{quote}
It is during this time that we see the introduction of mathematical criteria for fairness identical to the mathematical criteria of modern day.
Unfortunately, this fairness movement largely disappeared by the end of the 1970s, as the different and sometimes competing notions of fairness left little room for clarity on when one notion of fairness may be preferable to another.
\end{quote}
On the other hand, we find a similar warning in the failure of rule-based systems to deliver intelligent behavior in the 1980s, leading the whole field of AI into its second ``winter''~\citep{winter}:
\begin{quote}
  Eventually, the Advanced Research Projects Agency (ARPA), the research arm of the U.S.~Defense Department (later renamed to DARPA) and the primary funder of AI research and development, cut its funding of AI researchers because they had failed to deliver on most of their promises.
  At the time, the dominating form of creating software was still rule-based programming, in which developers explicitly specify all rules that define the behavior of a computer program.
\end{quote}
If crafting a complete set of rules describing ``what makes a cat a cat'' for a computer vision in the 80s system proved dauntingly difficult, why would this be now a correct strategy to define an intrinsically subjective concept such as fairness?

What could be an alternative to rule-based strategies to define and measure fairness?
We find hints in other fields of engineering tasked with measuring equally abstract concepts.
Some examples are measuring \emph{quality} in quality assurance~\citep{iso9000}, \emph{security} in penetration testing~\citep{arkin2005software}, \emph{reliability} in reliability engineering~\citep{o2012practical}, \emph{safety} in systems safety engineering~\citep{leveson2016engineering}, or \emph{correctness} in software unit testing~\citep{khorikov2020unit}.
In all of these instances, the measurement of some nuanced property about a complex system follows the same strategy:
by subjecting the system to inputs covering a wide range of experimental conditions, measuring the output of the system, and comparing the output to some expected behavior.
In the realm of machine learning, the research field concerned with evaluating the performance and robustness of classification systems under novel circumstances is the one of \emph{domain generalization}~\citep{gulrajani2020search}.

Curiously, the literature in domain generalization has a history of using metrics such as independence, separation, and sufficiency as regularizers to improve out-domain performance~\citep{federici2021information}, \citep[Table 1]{creager2021environment}.
In particular, independence is commonly implemented by domain-adversarial training of classifiers~\citep{ganin2016domain}, separation is similar but conditioned on the observed label~\citep{li2018deep}, and methods such as Invariant Risk Minimization~\citep[IRM]{arjovsky2019invariant, wald2021calibration} follow the sufficiency principle.

As it happens with fairness, there is no silver bullet to address all possible domain generalization problems.
Choosing the appropriate domain generalization regularizer is a task-dependent trouble that depends on the causal structure governing the data generating processes~\citep{gulrajani2020search, federici2021information}.
Unfortunately, these causal structures are never known in practice~\citep{peters2017elements}.
No matter what regularizer we choose, researchers in domain generalization always end up measuring the distance to their end goal the same way---by the ability of the resulting classifier to generalize across a different but related ``out-domain'' test distribution of examples relevant to the task at hand~\citep{gulrajani2020search}.

We posit that each of the claims about unfairness in the catalog from Section~\ref{sec:catalog} bears a hint about a \emph{failing to generalize out-domain}.
Thus, we propose to use the same ``generalization test'' pioneered in domain generalization literature to measure the fairness of a classification system.
This view shifts our focus from shaping fairness metrics to curating the examples over which these metrics are computed.
In short, we propose to measure their ability to generalize across \emph{multiple} and \emph{societally-relevant} collections of test examples.
This is a strategy that will allow (I) defining fairness flexibly on a task-per-task basis, (II) empowering all stakeholders to express their interests by proposing out-domain data, and (III) a less manipulable definition of fairness, as improving out-domain performance is challenging when out-domain data are not incorporated into the training set.

\section{The framework of stress tests}
\label{sec:stress}

We now introduce the framework of \emph{stress testing} to measure the fairness of machine learning systems.
Simply put, we measure fairness as the ability to generalize across multiple, societally-relevant distributions of examples collected for the task at hand.
We call each of these distributions one \emph{stress test}, and see it as one ``unit test'' probing the performance of the classifier across a collection of examples reflecting the interests of one or more of the stakeholders involved in the learning task.
A machine passes a stress test if it matches or exceeds a metric value (pre-defined in the metadata of the stress test) and fails it otherwise.
We say that the system passes a collection of stress tests $\mathcal{S}$ if it passes every stress test $S \in \mathcal{S}$ in the collection.
In that case, we say that the system is \emph{fair with respect to $\mathcal{S}$}.
Model cards~\citep{mitchell2019model} are helpful here, since they can reference the stress tests that the model passes or fails.

By way of introduction, let us summarize a quick tour of the stress testing pipeline, illustrated in Figure~\ref{fig:mainfig}.
Foremost, we encourage every stakeholder in the learning task to collect examples to illustrate their---possibly competing and even incompatible---interests, and to express their view on what it means for the system to be fair.
Stakeholders include data collectors, model trainers, ethics teams, activists, journalists, expert witnesses, groups, and individuals.
System creators subject the machine to each of the stress tests, and produce two deliverables to the stakeholders.
On the one hand, each stress test returns a ``pass'' or ``fail'' signal, depending on whether the system falls short or exceeds a predefined metric value (normally, classification accuracy).
On the other hand, the system returns the list of predictions made on the stress tests, which can be digitally signed (as we will see Section~\ref{sec:cert}) to hold system providers up to a degree of accountability.
Based on these deliverables, the different stakeholders discuss how to improve the fairness of the machine learning system.
Some avenues include upgrading the training data and stress tests, bettering the system architecture, limiting the scope of the problem, or even questioning if the task belongs to the realm of machine learning at all.
Note that knowledge about sensitive attributes is not necessary throughout the stress testing pipeline.

The rest of this section deepens our understanding of the main advantages provided by the framework of stress testing.
These include (I) flexibility, (II) subjectivity, and (III) robustness.
We also include a discussion on the uses of stress tests for model selection, some of the limitations of the framework, and connections to prior work in the fairness literature.

\subsection{Flexibility}

Stress tests shift focus from growing a catalog of fairness metrics to curating the distributions of examples under which we compute these metrics.
Moving from a rule-based to a data-based perspective opens the door to a more flexible, non-parametric paradigm to measure fairness in complex machine learning systems.
The framework of stress testing does not oppose but complements fairness metrics:
while we focus on classification performance, stress tests can rely on any fairness metric as long as it sheds light into the system under study.

Stress testing requires collecting multiple, dedicated distributions of examples to evaluate the fairness qualities of each new learning system.
While this is an admittedly large effort, it is a justified one based on the importance of providing fair services.
The dedicated collection of stress tests is also an avenue to incorporate contextual factors into the definition of fairness.
It is also an attempt to provide definitions of fairness that are honest and evolving, expressed by stress tests always subject to changing, enlarging, and improving.
Stress tests can also expire, as social norms change over time.

Stress tests can vary in size, and even contain a single example.
These could contain noteworthy cases revealing critical mistakes that would otherwise contribute a negligible percentage to average error~\citep{rostamzadeh2021thinking}.
Small, \emph{surgical} stress tests highlight the contribution of diversity in data, paramount to fairness~\citep{drosou2017diversity, jo2020lessons, chasalow2021representativeness}.
Mathematically, disintegrating one stress test of $n$ examples into $n$ stress tests of $1$ example amounts to requiring the correct classification across \emph{all examples}.
For instance, while attributing a non-human label to one single person has negligible impact on average error, we consider this single mistake enough to render the whole learning system unacceptable.

Stress tests define arbitrarily complex distributions of data.
In particular, stress go beyond ``sweeping over values of the sensitive attribute''.
They can involve general distribution shifts due to changing environmental factors in data collection, hidden confounders, unstable correlations, or difficult-to-define sensitive attributes (such as non-discrete or overlapping).
In many cases, access to sensitive attributes is not required to build such test: collecting the data only requires access to the variables that describe the target environment (e.g., location of defendants in the COMPAS example).
Furthermore, by selectively sampling specific regions of the feature space, stress testing allows the study of unfairness due to \emph{subgroup validity}, that is, features having different predictive power in different regions of their space~\citep{corbett2018measure}.

Finally, stress tests are an interpretable tool to study fairness.
In essence, they describe the input-output behaviour of the system on individual examples, which indeed matches the most common strategy to claim unfairness in the first place.

\subsection{Subjectivity}

Stress tests embrace the subjectivity involved in defining fairness, properly and explicitly accommodating all the stakeholders in the learning task.
More specifically, we invite every stakeholder to curate and evaluate one or multiple stress tests.
These stress tests are collections of examples that illustrate their---possibly competing and even incompatible---interests.
By amplifying the benefits observed in internal audits~\citep{raji2020closing}, the dynamics of stress testing dilute concentrations of power by distributing the ownership of the data involved in the training and evaluating learning systems.

At a higher level, we believe that evaluating a learning system on multiple distributions of examples is the richest language at our disposal to align machine and human incentives.
On the one hand, the more stress tests a classifier passes, the more confidence may human operators place on this system---after all, it is increasingly unlikely for the machine to find shortcuts to pass a growing number of diverse stress tests.
On the other hand, failing a stress test is a call to reconsider the available training data, improve the machine learning architecture, or redefining the scope of the task to provide a more honest service to users.

\subsection{Robustness}

The British economist Charles Goodhart once said ``when a metric becomes a target, it ceases to be a good metric''~\citep{strathern1997improving}.
This is particularly true for learning machines, given their incredible faculty to satisfy objectives in unexpected ways~\citep{geirhos2020shortcut}.
Examples of undesirable classification systems satisfying different metrics of fairness abound~\cite{corbett2018measure}.
Overall, it is more difficult for a machine (or for dishonest human operators) to cheat its way into passing a stress test of external data than satisfying a fixed metric.

The manipulability of stress testing decreases further when we place privacy constraints between stakeholders.
For instance, it is important to establish the necessary protocols to avoid the incorporation of stress examples into the training data.
Furthermore, it is possible to hide the data away from the model trainers, reporting only metric values over the stress test, or simply as pass/fail signal.
This would implement an interface akin to a Kaggle leaderboard~\citep{blum2015ladder} and allow the re-usability of the same stress data for multiple iterations.
Tools from differential privacy~\citep{dwork2015reusable} can further limit the information flow between stress data and system trainers.
When public, stress tests are also valuable contributions to the community, becoming datasets to measure \emph{out-domain} generalization and fairness.

\subsection{Aid to model selection}

As learning systems grow in complexity, it is common to obtain perfect models across the training distribution (in accuracy or some fairness metric).
Stress tests (examples collected from different distributions) are a tool to distinguish between zero-error models, which may have different levels of fairness out-domain.

More specifically, stress tests can criticize classifiers (even if these are perfect under the training distribution) by considering examples under two types of changes about the joint distribution $P(X, Y) = P(Y|X)P(X)$ describing the human cognitive process of transforming features into labels~\citep{arjovsky2019invariant}.
(We here consider that sensitive attributes, when available, are inside the overall feature vector $X$.)
On the one hand, changes in the labeling mechanism $P(Y|X)$ are often disregarded in the machine learning literature~\citep{peters2017elements, arjovsky2019invariant}.
In fairness, \citet{wachter2020bias} calls these shifts ``bias-transforming'' shifts.
Collecting examples under a different labeling mechanism is a tool to study label bias, described as one of the major roadblocks towards fair machine learning~\citep{corbett2018measure}.
In the COMPAS example, labeling recidivism when re-arresting happens (instead of actually re-offending) is an example of labeling bias~\citep{dressel2018accuracy}.
Here, a stress tests containing re-offending labels would illustrate the shortcomings of the implemented system.
On the other hand, changes in the distribution of inputs $P(X)$ consider the labeling mechanism invariant but explore the behavior of the learning across novel regions of examples.
In fairness, \citet{wachter2020bias} refers to these shifts as ``bias-preserving'' shifts.
In the COMPAS example, this would translate into stress tests collected in other cities, states, or countries.

Finally, stress tests may contain additional ``privileged'' features or labels not available to the learning system during training~\citep{lopez2015unifying}.
For instance, these may include sensitive attributes or better labels that were too expensive or even impossible to collect for training.
These additional features enable correlational analyses that can reveal types of unfairness otherwise unnoticed.

\subsection{Relation to prior work}
A narrow body of work mentions explicitly the link between fairness and out-domain generalization~\citep{federici2021information, creager2021environment}.
More implicitly, probing machine learning systems to different distributions of examples to assess their fairness is an increasingly popular practice~\citep{klare2015pushing, levi2015age, ImageNetRoulette, shankar2017no, de2019does, karkkainen2019fairface, karkkainen2019fairface, hazirbas2021towards, radford2021learning, schumann2021step, dollar}.

Stress tests are a type of audit studies~\citep{sandvig2014auditing, fairmlbook}. Engaging with such data-driven audits may open new ways for system developers to be accountable and take guided action with proper documentation~\citep{raji2020closing}.
Describing complex concepts---such as fairness---using rich input-output data instead of rigid rule-based systems is a natural progression that we have observed in other domains, such as in integration of mathematical expressions~\citep{lample2019deep}, protein folding in biology~\citep{senior2020improved}, and computer game playing~\citep{silver2017mastering}.
Other examples of engineering fields measuring nuanced properties via input-output behaviour include evaluating \emph{quality} in quality assurance~\citep{iso9000}, \emph{security} in penetration testing~\citep{arkin2005software}, \emph{reliability} in reliability engineering~\citep{o2012practical}, \emph{safety} in systems safety engineering~\citep{leveson2016engineering}, or \emph{correctness} in software unit testing~\citep{khorikov2020unit}.
In all of these fields, a nuanced property about a system (quality, security, safety, \emph{fairness}) is under scrutiny, and the set of exams carried out are (a) defined on a task-per-task basis, and (b) performed by multiple stakeholders.
In the future, we envision startup companies with the sole purpose of performing independent fairness audits of machine learning systems, where stress testing will likely play a major role.
These third-party independent consultants will assume an adversarial role (like a hacker performing a penetration test) against the systems they study, and their reputation will tie up to the amount of claims about unfairness that their audited systems star in the popular press.
The third-party fairness testers will have the option to publish their stress tests, which surely looks like a more convincing examination than passing a fixed, one-liner fairness metric.

Stress tests also relate to archives.
\citep{jo2020lessons} reviews literature in the archival and library sciences to improve our data collection practices by promoting ethics, representation, power, transparency, and consent.
The work argues for a separate, full-time area of expertise where all stakeholders have the power to build consensual, diverse, and correct datasets.
This is in line with our proposal to curate stress tests and developing the necessary guarantees to hold machine learning model providers accountable for their predictions.

\subsection{Limitations}

In some circumstances, stress testing is difficult or even impossible.
When stress testing a third-party system over the Internet, one may face legal consequences when querying the system indiscriminately~\citep{sandvig2014auditing}.
In some other situations, querying the system may be impossible for some of the stakeholders---for example, who can query a banking system that assists the granting or denial of loans?
Regulation and lawmaking on stress testing would aid these cases.
 
Stress tests shall contain only real-world examples: this excludes synthetic, adversarial or mislabeled examples~\citep{goodfellow2014explaining}.
Otherwise, bad actors could craft stress tests containing adversarial examples to produce any false pass/fail result of their liking.
One strategy to circumvent this limitation is to rely on oath taking in court.
Fabricated stress tests would in this case are similar to a false testimony or any other type of false evidence.

Another important question is the one on how to aggregate multiple pass/fail stress test results into a single ``fairness score'' describing the overall quality of a system.
We purposefully sidestep this problem, as we believe summarizing multiple stakeholders into a single number takes away from a fruitful discussion where different perspectives signal different problems that are difficult to compare.
While preventing a simplistic aggregation, the ongoing discussion between stakeholders is the main tool to improve the fairness of the system, as well as detecting and avoiding unrealistic, invalid or fake stress tests.
In particular, we propose that model cards~\citep{mitchell2019model} list the results of each stress test separately in order to provide the most comprehensive view of fairness possible.
In return, to inform system designers about how to improve their models, stress tests designers may use data sheets~\citep{gebru2021datasheets} to document their data while keeping the examples private.

The recommendations above apply to scenarios where system designers and stress testers share a common goal of making the system perform well for everyone.
In scenarios where the application domain is covered by an anti-discrimination laws, such as credit scoring or hiring, we can envision situations where failing a stress test could be used as a form of evidence of unlawful discrimination.
In that case, it is necessary to establish a relationship between the predictions given by a provider and the particular version (values of the model's parameters) of the system that produced such results.
Without such link, it would be impossible to enforce any sort of prediction accountability, as the provider could update the model \emph{ex post facto} any unfairness claim.
Given the importance of these use cases of stress tests, we outline a solution based on cryptography in the next section.

\section{Signed predictions}
\label{sec:cert}

It would be difficult to ensure the impact of stress testing without the ability to hold the system provider accountable for its predictions.
Lacking a mechanism to tie predictions to a particular learning system would allow the provider to secretly update the model after suffering a claim about unfairness, thus circumventing any accountability on the impact of its predictions on individuals and social groups.
In words of~\citet{wachter2020bias}, ``system controllers may, for example, limit the availability of relevant evidence to protect their intellectual property or avoid litigation''.

Our idea to enable prediction accountability is a variation of digital signatures~\citep{goldwasser1988digital}, which are mathematical protocols to verify the provenance of digital messages.
The goal here is in digitally signing predictions to link them to a particular snapshot of a machine learning system.
Digital signatures enable the accountability of all predictions, including those about the training data, all stress tests, and individual test examples.

Our starting point is to construct cryptographically-secure hashes~\citep{dworkin2015sha} of the model (weights), training data, and stress tests.
Since these three are objects stored up to permutations, we should document their ordering to reproduce their hashes when necessary.
For increased security, we can add a stream of random numbers to each object before hashing.
These random numbers, also known as ``salt'', must be securely stored~\citep{stinson2005cryptography}.
In the sequel, we denote these hashes by \hash{model}, \hash{training\_data} and \hash{stress\_k}, for some stress test $S \in \mathcal{S}$.

Next, we provide \hash{model} as a password to a secure password-based key derivation function such as \texttt{scrypt}~\citep{percival2009stronger}.
From this process we obtain a private/public pair, denoted by \private{} and \public.
The system provider makes \public{} accessible to all stakeholders.
At this point, the system provider may sign its prediction $r_i$ about some example $x_i$ by returning
\begin{center}
  \texttt{r\_i, signature := encrypt(concat(hash(x\_i), hash(r\_i)), model\_private)}.
\end{center}
The user may verify the provenance of \texttt{prediction} on their end by evaluating the equality
\begin{center}
  \texttt{decrypt(signature, model\_public) == concat(hash(x\_i), hash(r\_i))}
\end{center}
A similar protocol certifies that the machine learning model produces a fixed set of predictions over an entire dataset.
Of particular interest is to certify passing a particular stress test $S$ by publishing
\begin{center}
  \texttt{encrypt(concat(hash(x\^{}S\_i), hash(r\^{}S\_i)), model\_private),}
\end{center}
for all examples $(x^S_i, a^S_i, y^S_i)$ and prediction scores $r^S_i$ associated to the stress test $S$.
Note that it is not necessary to publish the data itself, reserving direct access to the examples for situations guaranteeing confidentiality agreements, such as in court.
Finally, stakeholders can sign stress tests to trace back and eliminate racist, fake, or mislabeled examples.

The previous scheme faces one main challenge.
Even in the case where private keys are properly secured, the system provider could employ different models to (i) compute the public/private key pair, and (ii) signing predictions.
This limits accountability, since predictions could arise from a new version of the classification system (fair, updated under the hood), yet signed with a private/public key pair associated to an old (unfair) snapshot.
One solution is to rely on third-party independent researchers, akin to expert witnesses in judicial processes.
Independent researchers could, under oath taking and the appropriate confidentiality agreements, reproduce the (i) key generation, and (ii) prediction processes to verify that the signatures provided to the user correspond to the appropriate model.
In such a way, machine learning models become the new ``crime's weapon''.

There are other considerations that can affect the correct signing of predictions.
These involve stochastic prediction processes, including the randomized augmentation of data, test-time adaptation of the classifier, reliance of the system on querying external data that changes over time, and any non-deterministic or parallelized operations.
Most of these issues are addressable by storing the state of all pseudo-random number generators as part of the signed data.
But, \emph{full accountability is possible only under full reproducibility}.

\section{Conclusion}
\label{sec:conclusion}

With the skyrocketing adoption of machine learning in societally-critical applications, measuring the fairness properties of automated decision systems is a chief subject of research.
Fairness is an incredibly nuanced concept to define, depending highly on context and the diverse perspectives of stakeholders.
Yet, our research community places an excessive trust on measuring fairness with rigid rules that are (I) encapsulated in mathematical one-liners, (II) do not offer enough opportunities for different stakeholders to voice their interests, and (III) are easy to manipulate and satisfy by undesirable classification systems.

Inspired by research literature in domain generalization, in this work we have advocated a transition from a parametric rule-based system into a non-parametric data-based system, where we shift our focus from shaping an ever-growing catalog of metrics into curating the distributions of examples over which these metrics are computed.
This is inline with fields of engineering that measure a difficult-to-define concept in terms of the input-output behavior of a system (quality assurance, computer security), and more novel strategy transitions from rules to data (such as numerical integration or protein folding).

The proposed framework of stress tests invites all stakeholders to curate societally-relevant distributions of examples that illustrate---their possibly competing and even incompatible---interests.
We believe that every claim about fairness should be immediately followed by the tagline: ``\emph{Fair under what examples, and collected by whom?}''.
The machine passes or fails each stress test (by falling short or exceeding some pre-defined metric value), informing a discussion between the stakeholders on how to improve the system.
This protocol enables (I) a definition of fairness that is context dependent and conditional to the inspected stress tests, (II) the proper involvement of all the stakeholders and their diverse perspectives, and (III) a less manipulable way to measure fairness, since system creators cannot cheat their way into reducing the test error over external stress data.

We believe machine learning predictions shape peoples lives and they will increasingly be seen as a form of authority.
Because of this, we closed our exposition by proposing a scheme based on asymmetric cryptography to guarantee a degree of accountability from system providers.
Overall, we believe that stress testing, by describing the behaviour of learning systems using multiple distributions of examples, is a rich language to describe, interpret, and reduce the ever-present gap between the human understanding of a task and the machine's shortcut to a solution.
We hope that the present work serves as a small step towards shrinking this gap, and as a new complementary tool towards implementing a fairer machine learning practice.

\section*{Acknowledgements}

We would like to thank deeply Nisha Deo for her tireless support throughout this project.

\bibliographystyle{ACM-Reference-Format}
\bibliography{bibliography}

\end{document}